\documentclass[10pt,twocolumn,letterpaper]{article}

\usepackage{cvpr}
\usepackage{times}
\usepackage{epsfig}
\usepackage{graphicx}
\usepackage{amsmath}
\usepackage{amssymb}

\usepackage{xspace}
\usepackage{color}
\usepackage{booktabs}
\usepackage{subcaption}
\usepackage{tikz}
\usetikzlibrary{positioning}
\usetikzlibrary{arrows.meta}
\usetikzlibrary{calc}
\tikzset{}

\usepackage{pgfplots}
\pgfplotsset{width=2.5cm,compat=newest}

\makeatletter
\renewcommand{\paragraph}{%
  \@startsection{paragraph}{4}%
  {\z@}{0.5ex \@plus 1ex \@minus .2ex}{-1em}%
  {\normalfont\normalsize\bfseries}%
}
\makeatother

\usepackage[pagebackref=true,breaklinks=true,letterpaper=true,colorlinks,bookmarks=false]{hyperref}

\newlength\tikzfigwidth
\newlength\tikzfigheight

\def\realspace{\mathbb{R}} 

\newcommand{\mat}[1]{\ensuremath{\mathbf{#1}}}
 \renewcommand{\vec}[1]{\ensuremath{\mathbf{#1}}}
\def\transp{\top} 

\newcommand{\eye}{\mat{I}} 

\newcommand{\nullspace}[1]{\ensuremath{\mathcal{N}\left(#1\right)}} 

\DeclareMathOperator*{\argmin}{arg\,min} 
\renewcommand{\det}[1]{\mbox{det}\left( #1 \right)} 

\newcommand{\chis}[2][non]{\ensuremath{\chi^2 \ifthenelse{\equal{#1}{non}}{}{ \left(#1,#2\right)}}} 
\newcommand{\Gammaf}[1][non]{\ensuremath{\Gamma\ifthenelse{\equal{#1}{non}}{}{ \left( #1 \right)}}} 

\newcommand{\degree}[1][non]{\ensuremath{\ifthenelse{\equal{#1}{non}}{^\circ}{#1^\circ}}} 
\LetLtxMacro{\oldsqrt}{\sqrt}
\renewcommand{\sqrt}[1][\ ]{%
  \def\DHLindex{#1}\mathpalette\DHLhksqrt}
\def\DHLhksqrt#1#2{%
  \setbox0=\hbox{$#1\oldsqrt[\DHLindex]{#2\,}$}\dimen0=\ht0
  \advance\dimen0-0.2\ht0
  \setbox2=\hbox{\vrule height\ht0 depth -\dimen0}%
  {\box0\lower0.71pt\box2}}
\newcommand{\cardinality}[1]{\ensuremath{\left|#1\right|}} 

\newcommand{\loss}[1]{\ensuremath{\mathcal{L}\left(#1\right)}} 
\def\lossSym{\mathcal{L}}

\def\flow{\vec{x}}
\def\flowIn{x^{\textrm{in}}}
\def\flowOut{x^{\textrm{out}}}
\def\flowDet{x^{\textrm{det}}}
\def\flowLink{x^{\textrm{link}}}
\def\flowOpt{\flow^*}
\def\flowGt{\flow^{\textrm{gt}}}

\def\cost{\vec{c}}
\def\costIn{c^{\textrm{in}}}
\def\costOut{c^{\textrm{out}}}
\def\costDet{c^{\textrm{det}}}
\def\costLink{c^{\textrm{link}}}

\def\costUnary{\vec{c}^{\textrm{U}}}
\def\costPairwise{c^{\textrm{P}}}

\def\costFin{\vec{f}}
\def\costFpa{\mathbf{\Theta}}

\def\Nvars{M}

\def\opttemp{t}

\def\det{\vec{d}} 

\def\bboxIdx{i} 
\def\bboxIdxAlt{j} 
\def\bboxN{K} 

\def\bbox{\vec{b}(\det)} 
\newcommand{\detBox}[1]{\ensuremath{\vec{b}(#1)}}
\def\bboxP{p(\det)} 
\newcommand{\detProb}[1]{\ensuremath{p(#1)}}
\def\bboxT{t(\det)}
\newcommand{\detFrame}[1]{\ensuremath{t(#1)}}

\def\traj{\mathcal{T}}
\def\trajIdx{k}
\def\trajSize{N_\trajIdx}

\def\winsize{W}

\cvprfinalcopy 

\def\cvprPaperID{3208} 

\ifcvprfinal\fi
\begin{document}

\title{Deep Network Flow for Multi-Object Tracking}

\author{Samuel Schulter \quad Paul Vernaza \quad Wongun Choi \quad Manmohan Chandraker \\
  NEC Laboratories America, Media Analytics Department \\
  Cupertino, CA, USA \\
  {\small\texttt{\{samuel,pvernaza,wongun,manu\}@nec-labs.com}}
}

\maketitle
\thispagestyle{empty}

\begin{abstract}
Data association problems are an important component of many computer vision applications, with multi-object tracking being one of the most prominent examples.
A typical approach to data association involves finding a graph matching or network flow that minimizes a sum of pairwise association costs, which are often either hand-crafted or learned as linear functions of fixed features.
In this work, we demonstrate that it is possible to learn features for network-flow-based data association via backpropagation, by expressing the optimum of a smoothed network flow problem as a differentiable function of the pairwise association costs.
We apply this approach to multi-object tracking with a network flow formulation.
Our experiments demonstrate that we are able to successfully learn all cost functions for the association problem in an end-to-end fashion, which outperform hand-crafted costs in all settings.
The integration and combination of various sources of inputs becomes easy and the cost functions can be learned entirely from data, alleviating tedious hand-designing of costs.

\end{abstract}

\section{Introduction}
\label{sec:introduction}

Multi-object tracking (MOT) is the task of predicting the trajectories of all object instances in a video sequence.
MOT is challenging due to occlusions, fast moving objects or moving camera platforms, but it is an essential module in many applications like action recognition, surveillance or autonomous driving.
The currently predominant approach to MOT is tracking-by-detection~\cite{sam:Berclaz11a,sam:Breitenstein09a,sam:Choi15a,sam:Ess09a,sam:Kuo10a,sam:Milan14a,sam:Possegger14a}, where, in a first step, object detectors like~\cite{sam:Felzenszwalb10a,sam:Ren15a,sam:Wang13a} provide potential locations of the objects of interest in the form of bounding boxes.
Then, the task of multi-object tracking translates into a data association problem where the bounding boxes are assigned to trajectories that describe the path of individual object instances over time.

Bipartite graph matching~\cite{sam:Kuhn55a,sam:Munkres57a} is often employed in on-line approaches to assign bounding boxes in the current frame to existing trajectories~\cite{sam:Khan05a,sam:Oh09a,sam:Okuma04a,sam:Xiang15a}.
Off-line methods can be elegantly formulated in a network flow framework to solve the association problem including birth and death of trajectories~\cite{sam:Leal-Taixe16a,sam:Leal-Taixe11a,sam:Zhang08a}.
Section~\ref{sec:related_work} gives more examples.
All these association problems can be solved in a linear programming (LP) framework, where the constraints are given by the problem.
The interplay of all variables in the LP, and consequently their costs, determines the success of the tracking approach.
Hence, designing good cost functions is crucial.
Although cost functions are hand-crafted in most prior work, there exist approaches for learning costs from data.
However, they either do not treat the problem as a whole and only optimize parts of the costs~\cite{sam:Leal-Taixe16a,sam:Li09a,sam:Xiang15a,sam:Zhang08a} or are limited to linear cost functions~\cite{sam:Wang15a,sam:Wang16a}.

We propose a novel formulation that allows for learning \emph{arbitrary parameterized cost functions} for \emph{all variables} of the association problem in an end-to-end fashion, \ie, from input data to the \emph{solution of the LP}.
By smoothing the LP, bi-level optimization~\cite{sam:Bracken73,sam:Colson07a} enables learning of all the parameters of the cost functions such as to minimize a loss that is defined on the solution of the association problem, see Section~\ref{subsec:method_learning_cost_functions_from_data}.
The main benefit of this formulation is its flexibility, general applicability to many problems and the avoidance of tedious hand-crafting of cost functions.
Our approach is not limited to log-linear models (\cf,~\cite{sam:Wang15a}) but can take full advantage of any differentiable parameterized function, \eg, neural networks, to predict costs.
Indeed, our formulation can be integrated into any deep learning framework as one particular layer that solves a linear program in the forward pass and back-propagates gradients \wrt the costs through its solution (see Figure~\ref{fig:mot_inference}).

While our approach is general and can be used for many association problems, we explore its use for multi-object tracking with a network flow formulation (see Sections~\ref{subsec:method_network_flow_formulation} and \ref{subsec:method_tracking_model}).
We empirically demonstrate on public data sets~\cite{sam:Geiger12a,sam:Leal-Taixe15a,sam:Milan16a} that:
(i) Our approach enables end-to-end learning of cost functions for the network flow problem.
(ii) Integrating different types of input sources like bounding box information, temporal differences, appearance and motion features becomes easy and all model parameters can be learned jointly.
(iii) The end-to-end learned cost functions outperform hand-crafted functions without the need to hand-tune parameters.
(iv) We achieve encouraging results with appearance features, which suggests potential benefits from end-to-end integration of deep object detection and tracking, as enabled by our formulation.


\section{Related Work}
\label{sec:related_work}

\paragraph{Association problems in MOT:}
Recent works on multi-object tracking (MOT) mostly follow the tracking-by-detection paradigm~\cite{sam:Berclaz11a,sam:Breitenstein09a,sam:Choi15a,sam:Ess09a,sam:Kuo10a,sam:Milan14a,sam:Possegger14a}, where objects are first detected in each frame and then associated over time to form trajectories for each object instance.
On-line methods like~\cite{sam:Breitenstein11a,sam:Choi13a,sam:Ess09a,sam:Pellegrini09a,sam:Possegger14a} associate detections of the incoming frame immediately to existing trajectories and are thus appropriate for real-time applications\footnote{In this context, real-time refers to a causal system.}.
Trajectories are typically treated as state-space models like Kalman~\cite{sam:Kalman60} or particle filters~\cite{sam:Gordon93a}.
The association to bounding boxes in the current frame is often formulated as bipartite graph matching and solved via the Hungarian algorithm~\cite{sam:Kuhn55a,sam:Munkres57a}.
While on-line methods only have access to the past and current observations, off-line (or batch) approaches~\cite{sam:Berclaz11a,sam:Butt13a,sam:Huang08a,sam:Milan12a,sam:Pirsiavash11a,sam:Zhang08a} also consider future frames or even the whole sequence at once.
Although not applicable for real-time applications, the advantage of batch methods is the temporal context allowing for more robust and non-greedy predictions.
An elegant solution to assign trajectories to detections is the network flow formulation~\cite{sam:Zhang08a} (see Section~\ref{subsec:method_network_flow_formulation} for details).
Both of these association models can be formulated as linear program.

\paragraph{Cost functions:}
Independent of the type of association model, a proper choice of the cost function is crucial for good tracking performance.
Many works rely on carefully designed but hand-crafted functions.
 For instance, \cite{sam:Leal-Taixe11a,sam:Milan14a,sam:Possegger14a} only rely on detection confidences and spatial (\ie., bounding box differences) and temporal distances.
Zhang~\etal~\cite{sam:Zhang08a} and Zamir~\etal~\cite{sam:Zamir12a} include appearance information via color histograms.
Other works explicitly learn affinity metrics, which are then used in their tracking formulation.
For instance, Li~\etal~\cite{sam:Li09a} build upon a hierarchical association approach where increasingly longer tracklets are combined into trajectories.
Affinities between tracklets are learned via a boosting formulation from various hand-crafted inputs including length of trajectories and color histograms.
This approach is extended in \cite{sam:Kuo10a} by learning affinities on-line for each sequence.
Similarly, Bae and Yoon~\cite{sam:Bae14a} learn affinities on-line with a variant of linear discriminant analysis.
Song~\etal~\cite{sam:Song08a} train appearance models on-line for individual trajectories when they are isolated, which can then be used to disambiguate from other trajectories in difficult situations like occlusions or interactions.
Leal-Taix\'{e}~\etal~\cite{sam:Leal-Taixe16a} train a Siamese neural network to compare the appearance (raw RGB patches) of two detections and combine this with spatial and temporal differences in a boosting framework.
These pair-wise costs are used in a network flow formulation similar to~\cite{sam:Leal-Taixe11a}.
In contrast to our approach, none of these methods consider the actual inference model during the learning phase but rely on surrogate loss functions for parts of the tracking costs.

\paragraph{Integrating inference into learning:}
Similar to our approach, there have been recent works that also include the full inference model in the training phase.
In particular, structured SVMs~\cite{sam:Tsochantaridis05a} have recently been used in the tracking context to learn costs for bipartite graph matching in an on-line tracker~\cite{sam:Kim12a}, a divide-and-conquer tracking strategy~\cite{sam:Solera15a} and a joint graphical model for activity recognition and tracking~\cite{sam:Choi12a}.
In a similar fashion, \cite{sam:Wang15a} present a formulation to jointly learn all costs in a network flow graph with a structured SVM, which is the closest work to ours.
It shows that properly learning cost functions for a relatively simple model can compete with complex tracking approaches.
However, the employed structured SVM limits the cost functions to a linear parameterization.
In contrast, our approach relies on bi-level optimization~\cite{sam:Bracken73,sam:Colson07a} and is more flexible, allowing for non-linear (differentiable) cost functions like neural networks.
Bi-level optimization has also been used recently to learn costs of graphical models, \eg, for segmentation~\cite{sam:Ranftl14a} or depth map restoration~\cite{sam:Riegler15a,sam:Riegler16a}.


\section{Deep Network Flows for Tracking}
\label{sec:method}
We demonstrate our end-to-end formulation for association problems with the example of network flows for multi-object tracking.
In particular, we consider a tracking-by-detection framework, where potential detections $\det$ in every frame $t$ of a video sequence are given.
Each detection consists of a bounding box $\bbox$ describing the spatial location, a detection probability $\bboxP$ and a frame number $\bboxT$.
For each detection, the tracking algorithm needs to either associate it with an object trajectory $\traj_\trajIdx$ or reject it.
A trajectory is defined as a set of detections belonging to the same object, \ie, $\traj_\trajIdx = \{\det_\trajIdx^1, \ldots, \det_\trajIdx^{\trajSize}\}$, where $\trajSize$ defines the size of the trajectory.
Only bounding boxes from different frames can belong to the same trajectory.
The number of trajectories $\cardinality{\traj}$ is unknown and needs to be inferred as well.

In this work, we focus on the network flow formulation from Zhang~\etal~\cite{sam:Zhang08a} to solve the association problem.
It is a popular choice~\cite{sam:Leal-Taixe16a,sam:Leal-Taixe11a,sam:Lenz15a,sam:Wang15a} that works well in practice and can be solved via linear programming (LP).
Note that bipartite graph matching, which is typically used for on-line trackers, can also be formulated as a network flow, making our learning approach equally applicable.

\subsection{Network Flow Formulation}
\label{subsec:method_network_flow_formulation}
\begin{figure}\centering
\input{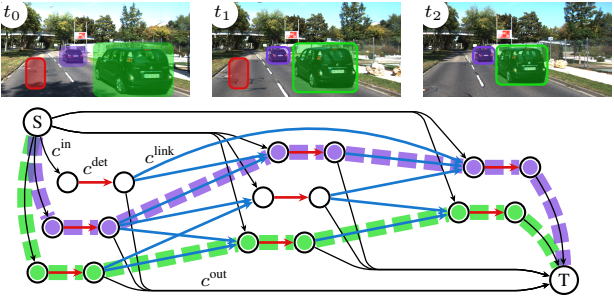}
\caption{
A network flow graph for tracking $3$ frames~\cite{sam:Zhang08a}.
Each pair of nodes corresponds to a detection.
The different solid edges are explained in the text, the thick dashed lines illustrate the solution of the network flow.
}
\label{fig:network_flow_graph_example}
\end{figure}

We present the formulation of the directed network flow graph with an example illustrated in Figure~\ref{fig:network_flow_graph_example}.
Each detection $\det_\bboxIdx$ is represented with two nodes connected by an edge (red).
This edge is assigned the flow variable $\flowDet_\bboxIdx$.
To be able to associate two detections, meaning they belong to the same trajectory $\traj$, directed edges (blue) from all $\det_\bboxIdx$ (second node) to all $\det_\bboxIdxAlt$ (first node) are added to the graph if $\detFrame{\det_\bboxIdx} < \detFrame{\det_\bboxIdxAlt}$ and $|\detFrame{\det_\bboxIdx} - \detFrame{\det_\bboxIdxAlt}| < \tau_t$.
Each of these edges is assigned a flow variable $\flowLink_{\bboxIdx,\bboxIdxAlt}$.
Having edges over multiple frames allows for handling occlusions or missed detections.
To reduce the size of the graph, we drop edges between detections that are spatially far apart.
This choice relies on the smoothness assumption of objects in videos and does not hurt performance but reduces inference time.
In order to handle birth and death of trajectories, two special nodes are added to the graph.
A source node (S) is connected with the first node of each detection $\det_\bboxIdx$ with an edge (black) that is assigned the flow variable $\flowIn_\bboxIdx$.
Similarly, the second node of each detection is connected with a sink node (T) and the corresponding edge (black) is assigned the variable $\flowOut_\bboxIdx$.

Each variable in the graph is associated with a cost.
For each of the four variable types we define the corresponding cost, \ie, $\costIn$, $\costOut$, $\costDet$ and $\costLink$.
For ease of explanation later, we differentiate between unary costs $\costUnary$ ($\costIn$, $\costOut$ and $\costDet$) and pairwise costs $\costPairwise$ ($\costLink$).
Finding the globally optimal minimum cost flow can be formulated as the linear program
\begin{equation}
\begin{aligned}
  \flow^* =& \; \argmin_{\flow} \cost^\transp \flow \\
  \text{s.t.} & \;\; \mat{A} \flow \le \vec{b}, \, \mat{C} \flow = \vec{0},
\end{aligned}
\label{eq:inference}
\end{equation}
where $\flow \in \realspace^{\Nvars}$ and $\cost \in \realspace^{\Nvars}$ are the concatenations of all flow variables and costs, respectively, and $\Nvars$ is the problem dimension.
Note that we already relaxed the actual integer constraint on $\flow$ with box constraints $0 \le \flow \le 1$, modeled by $\mat{A} = [\eye, -\eye]^\transp \in \realspace^{2\Nvars \times \Nvars}$ and $\vec{b} = [\vec{1}, \vec{0}]^\transp \in \realspace^{2\Nvars}$ in~\eqref{eq:inference}.
The flow conservation constraints, $\flowIn_\bboxIdx + \sum_\bboxIdxAlt \flowLink_{\bboxIdxAlt\bboxIdx} = \flowDet_\bboxIdx$ and $\flowOut_\bboxIdx + \sum_\bboxIdxAlt \flowLink_{\bboxIdx\bboxIdxAlt} = \flowDet_\bboxIdx \; \forall \bboxIdx$, are modeled with $\mat{C} \in \realspace^{2\bboxN \times \Nvars}$, where $\bboxN$ is the number of detections.
The thick dashed lines in Figure~\ref{fig:network_flow_graph_example} illustrate $\flow^*$.

The most crucial part in this formulation is to find proper costs $\cost$ that model the interplay between birth, existence, death and association of detections.
The final tracking result mainly depends on the choice of $\cost$.

\subsection{End-to-end Learning of Cost Functions}
\label{subsec:method_learning_cost_functions_from_data}
The main contribution of this paper is a flexible framework to learn functions that predict the costs of \emph{all} variables in the network flow graph.
Learning can be done end-to-end, \ie, from the input data all the way to the solution of the network flow problem.
To do so, we replace the constant costs $\cost$ in Equation~\eqref{eq:inference} with parameterized cost functions $\cost(\costFin, \costFpa)$, where $\costFpa$ are the parameters to be learned and $\costFin$ is the input data.
For the task of MOT, the input data typically are bounding boxes, detection scores, images features, or more specialized and effective features like ALFD~\cite{sam:Choi15a}.

Given a set of ground truth network flow solutions $\flowGt$ of a tracking sequence (we show how to define ground truth in Section~\ref{subsec:method_learning_cost_functions_from_data_gt_and_loss}) and the corresponding input data $\costFin$, we want to learn the parameters $\costFpa$ such that the network flow solution minimizes some loss function.
This can be formulated as the bi-level optimization problem
\begin{equation}
\begin{aligned}
  \argmin_{\Theta} & \; \loss{\flow^{\textrm{gt}}, \flow^*} \\
  \text{s.t.}      & \;\; \flow^* = \argmin_{\flow} \cost(\costFin, \Theta)^\transp \flow \\
                   & \;\; \mat{A} \flow \le \vec{b}, \, \mat{C} \flow = \vec{0},
\end{aligned}
\label{eq:bilevel_problem_raw}
\end{equation}
which tries to minimize the loss function $\lossSym$ (upper level problem) \wrt the solution of another optimization problem (lower level problem), which is the network flow in our case, \ie, the inference of the tracker.
To compute gradients of the loss function \wrt the parameters $\costFpa$ we require a smooth lower level problem.
The box constraints, however, render it non-smooth.

\subsubsection{Smoothing the lower level problem}
\label{subsubsec:method_learning_cost_functions_from_data_smoothinglowerlevel}
The box constraints in \eqref{eq:inference} and \eqref{eq:bilevel_problem_raw} can be approximated via log-barriers~\cite{sam:Boyd04}.
The inference problem then becomes
\begin{equation}
\begin{aligned}
  \flow^* =& \; \argmin_{\flow} \opttemp \cdot \cost(\costFin, \Theta)^\transp \flow - \sum_{i=1}^{2\Nvars} \log(b_i - \vec{a}_i^\transp\flow) \\
  \text{s.t.} & \;\; \mat{C} \flow = \vec{0},
\end{aligned}
\end{equation}
where $\opttemp$ is a temperature parameter (defining the accuracy of the approximation) and $\vec{a}_i^\transp$ are rows of $\mat{A}$.
Moreover, we can get rid of the linear equality constraints with a change of basis $\flow = \flow(\vec{z}) = \flow_0 + \mat{B}\vec{z}$, where $\mat{C}\flow_0=\vec{0}$ and $\mat{B} = \nullspace{\mat{C}}$, \ie, the null space of $\mat{C}$, making our objective unconstrained in $\vec{z}$ ($\mat{C}\flow=\mat{C}\flow_0 + \mat{C}\mat{B}\vec{z}=\mat{C}\flow_0=\vec{0} = \textrm{True} \; \forall \vec{z}$).
This results in the following unconstrained and smooth lower level problem
\begin{equation}
\argmin_{\vec{z}} \opttemp \cdot \cost(\costFin, \Theta)^\transp \flow(\vec{z}) + P(\flow(\vec{z})),
\label{eq:smoothed_lower_level}
\end{equation}
where $P(\flow) = -\sum_{i=1}^{2\Nvars}\log(b_i - \vec{a}_i^\transp\flow)$.

\subsubsection{Gradients with respect to costs}
\label{subsubsec:method_learning_cost_functions_from_data_gradients}
Given the smoothed lower level problem \eqref{eq:smoothed_lower_level}, we can define the final learning objective as
\begin{equation}
\begin{aligned}
  \argmin_{\costFpa} & \; \loss{\flow^{\textrm{gt}}, \flow(\vec{z}^*}) \\
  \text{s.t.}        & \; \vec{z}^* = \argmin_{\vec{z}} t \cdot \cost(\costFin, \costFpa)^\transp \flow(\vec{z}) + P(\flow(\vec{z})),
\end{aligned}
\label{eq:bilevel_final}
\end{equation}
which is now well-defined.
We are interested in computing the gradient of the loss $\lossSym$ \wrt the parameters $\costFpa$ of our cost function $\cost(\cdot, \costFpa)$.
It is sufficient to show $\frac{\partial \lossSym}{\partial \cost}$, as gradients for the parameters $\costFpa$ can be obtained via the chain rule assuming $\cost(\cdot; \costFpa)$ is differentiable \wrt $\costFpa$.

The basic idea for computing gradients of problem~\eqref{eq:bilevel_final} is to make use of implicit differentiation on the optimality condition of the lower level problem.
For an uncluttered notation, we drop all dependencies of functions in the following.
We define the desired gradient via chain rule as
\begin{equation}
\begin{aligned}
  \frac{\partial\lossSym}{\cost} &= \; \frac{\partial\vec{z}^*}{\partial\cost} \frac{\partial\flow}{\partial\vec{z}^*} \frac{\partial\lossSym}{\partial\flow} = \frac{\partial\vec{z}^*}{\partial\cost} \mat{B}^\transp \frac{\partial\lossSym}{\partial\flow}.
\end{aligned}
\end{equation}
We assume the loss function $\lossSym$ to be differentiable \wrt \flow.
To compute $\frac{\partial\vec{z}^*}{\partial\cost}$, we use the optimality condition of~\eqref{eq:smoothed_lower_level}
\begin{equation}
\begin{aligned}
  \vec{0} &= \frac{\partial}{\partial\vec{z}} \left[ \opttemp \cdot \cost^\transp \flow + P \right] \\
  &= \opttemp \cdot \frac{\partial\flow}{\partial\vec{z}} \cost + \frac{\partial\flow}{\partial\vec{z}} \frac{\partial P}{\partial\flow} = \opttemp \cdot \mat{B}^\transp\cost + \mat{B}^\transp \frac{\partial P}{\partial\flow}
\end{aligned}
\end{equation}
and differentiate \wrt $\cost$, which gives
\begin{equation}
\begin{aligned}
  \vec{0} &= \frac{\partial}{\partial\cost} \left[ \opttemp \cdot \mat{B}^\transp \cost \right] + \frac{\partial}{\partial\cost} \left[ \mat{B}^\transp \frac{\partial P}{\partial\flow} \right] \\
          &= \opttemp \cdot \mat{B} + \frac{\partial\vec{z}}{\partial\cost} \frac{\partial\flow}{\partial\vec{z}} \frac{\partial^2 P}{\partial\flow^2} \mat{B} = \opttemp \cdot \mat{B} + \frac{\partial\vec{z}}{\partial\cost} \mat{B}^\transp \frac{\partial^2 P}{\partial\flow^2} \mat{B}
\end{aligned}
\end{equation}
and which can be rearranged to
\begin{equation}
\begin{aligned}
    \frac{\partial\vec{z}}{\partial\cost} = -\opttemp \cdot \mat{B}  \left[ \mat{B}^\transp \frac{\partial^2 P}{\partial\flow^2} \mat{B} \right]^{-1}.
\end{aligned}
\end{equation}
The final derivative can then be written as
\begin{equation}
\begin{aligned}
    \frac{\partial\lossSym}{\cost} = -\opttemp \cdot \mat{B} \left[ \mat{B}^\transp \frac{\partial^2 P}{\partial\flow^2} \mat{B} \right]^{-1} \mat{B}^\transp \frac{\partial\lossSym}{\partial\flow}.
\end{aligned}
\label{eq:bilevel_final_gradients}
\end{equation}
To fully define \eqref{eq:bilevel_final_gradients}, we provide the second derivative of $P$ \wrt $\flow$, which is given as
\begin{equation}
\begin{aligned}
    \frac{\partial^2 P}{\partial\flow^2} = \frac{\partial^2 P}{\partial\flow\partial\flow^\transp} = \sum_{i=1}^{2\Nvars} \frac{1}{\left(b_i - \vec{a}_i^\transp\flow\right)^2} \cdot  \vec{a}_i \vec{a}_i^\transp.
\end{aligned}
\end{equation}
In the supplemental material we show that \eqref{eq:bilevel_final_gradients} is equivalent to a generic solution provided in~\cite{sam:Ochs15a} and that $\mat{B}^\transp \frac{\partial^2 P}{\partial\flow^2} \mat{B}$ is always invertible.

\begin{figure*}\centering
\input{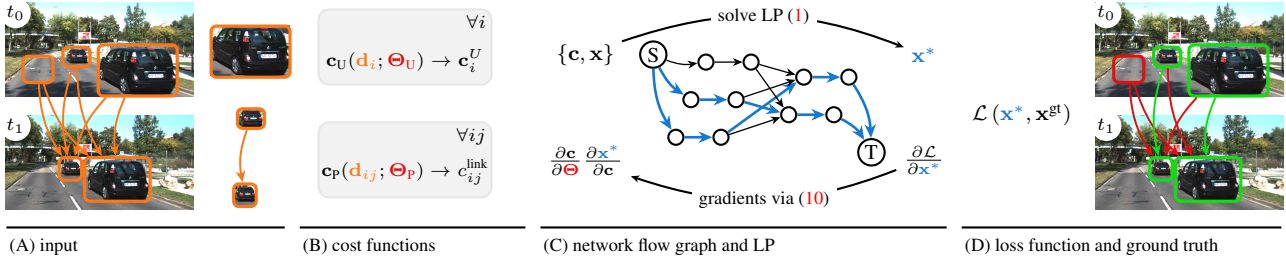}
\vspace{-0.2cm}
\caption{
During inference, two cost functions (B) predict unary and pair-wise costs based on features extracted from detections on the input frames (A).
The costs drive the network flow (C).
During training, a loss compares the solution $\flowOpt$ with ground truth $\flowGt$ to back-propagate gradients to the parameters $\costFpa$.
}
\label{fig:mot_inference}
\end{figure*}

\subsubsection{Discussion}
\label{subsubsec:method_learning_cost_functions_from_data_discussion}
Training requires to solve the smoothed linear program~\eqref{eq:smoothed_lower_level}, which can be done with any convex solver.
This is essentially one step in a path-following method with a fixed temperature $\opttemp$.
As suggested in~\cite{sam:Boyd04}, we set $\opttemp = \frac{\Nvars}{\epsilon}$, where $\epsilon$ is a hyper-parameter defining the approximation accuracy of the log barriers.
We tried different values for $\epsilon$ and also an annealing scheme, but the results seem insensitive to this choice.
We found $\epsilon = 0.1$ to work well in practice.

It is also important to note that our formulation is not limited to the task of MOT.
It can be employed for any application where it is desirable to learn costs functions from data for an association problem, or, more generally, for a linear program with the assumptions given in Section~\ref{subsubsec:method_learning_cost_functions_from_data_smoothinglowerlevel}.
Our formulation can also be interpreted as one particular layer in a neural network that solves a linear program.
The analogy between solving the smoothed linear program~\eqref{eq:smoothed_lower_level} and computing the gradients~\eqref{eq:bilevel_final_gradients} with the forward and backward pass of a layer in a neural network is illustrated in Figure~\ref{fig:mot_inference}.

\subsection{Defining ground truth and the loss function}
\label{subsec:method_learning_cost_functions_from_data_gt_and_loss}
To learn the parameters $\costFpa$ of the cost functions we need to compare the LP solution $\flowOpt$ with the ground truth solution $\flowGt$ in a loss function $\lossSym$.
Basically, $\flowGt$ defines which edges in the network flow graph should be active ($x^{\textrm{gt}}_i = 1$) and inactive ($x^{\textrm{gt}}_i = 0$).
Training data needs to contain the ground truth bounding boxes (with target identities) and the detection bounding boxes.
The detections define the structure of the network flow graph (see Section~\ref{subsec:method_network_flow_formulation}).

To generate $\flowGt$, we first match each detection with ground truth boxes in each frame individually.
Similar to the evaluation of object detectors, we match the highest scoring detection having an intersection-over-union overlap larger $0.5$ to each ground truth bounding box.
This divides the set of detection into true and false positives and already defines the ground truth for $\flowDet$.
In order to provide ground truth for associations between detections, \ie, $\flowLink$, we iterate the frames sequentially and investigate all edges pointing forward in time for each detection.
We activate the edge that points to the closest true positive detection in time, which has the same target identity.
All other $\flowLink$ edges are set to $0$.
After all ground truth trajectories are identified, it is straightforward to set the ground truth of $\flowIn$ and $\flowOut$.

\begin{figure}\centering
\input{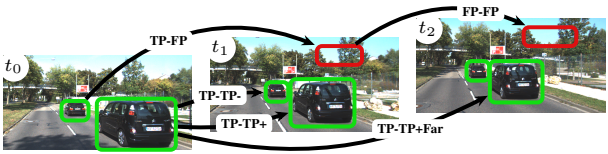}
\vspace{-0.3cm}
\caption{
An illustration of different types of links that emerge when computing the loss.
See text for more details on the different combinations of true (TP, green) and false positive (FP, red) detections.
}
\label{fig:mainpart_ground_truth_link_types}
\end{figure}

As already pointed out in~\cite{sam:Wang16a}, there exist different types of links that should be treated differently in the loss function.
There are edges $\flowLink$ between two false positives (FP-FP), between true and false positives (TP-FP), and between two true positives with the same (TP-TP+) or a different (TP-TP-) identity.
For (TP-TP+) links, we also differentiate between the shortest links for the trajectory and links that are longer (TP-TP+Far).
Edges associated with a single detection ($\flowIn$, $\flowDet$ and $\flowOut$) are either true (TP) or false positives (FP).
Figure~\ref{fig:mainpart_ground_truth_link_types} illustrates all these cases.
To trade-off the importance between these types, we define the following weighted loss function
\begin{equation}
\begin{aligned}
  \loss{\flow^*,\flowGt} =& \sum_{\kappa \in \{\textrm{in}, \textrm{det}, \textrm{out}\}} \sum_{\bboxIdx} \omega_i (x^{\kappa,*}_{\bboxIdx} - x^{\textrm{gt}}_\bboxIdx)^2 \\
                          & \; + \sum_{i,j \in \mathcal{E}} \omega_{ij} (x^{\textrm{link},*}_{\bboxIdx,\bboxIdxAlt} - x^{\textrm{gt}}_{\bboxIdx,\bboxIdxAlt})^2,
\end{aligned}
\end{equation}
where $\mathcal{E}$ is the set of all edges between detections $\bboxIdx$ and $\bboxIdxAlt$.
Note that the weights can be adjusted for each variable separately.
The default value for the weights is $1$, but we can adjust them to incorporate three intuitions about the loss.
(i) Ambiguous edges: Detections of an (FP-FP) link may describe a consistently tracked but wrong object.
Also, detections of a (TP-TP+Far) link are obviously very similar.
In both cases the ground truth variable is still inactive.
It may hurt the learning procedure if a wrong prediction is penalized too much for these cases.
Thus, we can set $\omega_{\bboxIdx,\bboxIdxAlt} = \omega_{\textrm{amb}} < 1$.
(ii) To influence the trade-off between precision and recall, we define the weight $\omega_{\textrm{pr}}$ for all edges involving a true positive detection.
Increasing $\omega_{\textrm{pr}}$ favors recall.
(iii) To emphasize associations, we additionally weight all $\flowLink$ variables with $\omega_{\textrm{link}}$.
If multiple of these cases are true for a single variable, we multiply the weights.

Finally, we note that~\cite{sam:Wang16a} uses a different weighting scheme and an $\ell_1$ loss.
We compare this definition with various weightings of our loss function in Section~\ref{subsec:experiments_weighting_the_loss_function}.

\subsection{Tracking model}
\label{subsec:method_tracking_model}
After the training phase, the above described network flow formulation can be readily applied for tracking.
One option is to batch process whole sequences at once, which, however, does not scale to long sequences.
Lenz~\etal~\cite{sam:Lenz15a} present a sophisticated approximation with bounded memory and computation costs.
As we focus on the learning phase in this paper, we opt for a simpler approach, which empirically gives similar results to batch processing but does not come with guarantees as in~\cite{sam:Lenz15a}.

We use a temporal sliding window of length $\winsize$ that breaks a video sequence into chunks.
We solve the LP problem for the frames inside the window, move it by $\Delta$ frames and solve the new LP problem, where $0 < \Delta < \winsize$ ensures a minimal overlap of the two solutions.
Each solution contains a separate set of trajectories, which we associate with bipartite graph matching to carry the object identity information over time.
The matching cost for each pair of trajectories is inversely proportional to the number of detections they share.
Unmatched trajectories get new identities.

In practice, we use maximal overlap, \ie, $\Delta = 1$, to ensure stable associations of trajectories between two LP solutions.
For each window, we output the detections of the middle frame, \ie, looking $\frac{\winsize}{2}$ frames into future and past, similar to~\cite{sam:Choi15a}.
Note that using detections from the latest frame as output enables on-line processing.


\section{Experiments}
\label{sec:experiments}
To evaluate the proposed tracking algorithm we use the publicly available benchmarks KITTI tracking~\cite{sam:Geiger12a}, MOT15~\cite{sam:Leal-Taixe15a} and MOT16~\cite{sam:Milan16a}.
The data sets provide training sets of $21$, $11$ and $7$ sequences, respectively, which are fully annotated.
As suggested in~\cite{sam:Geiger12a,sam:Leal-Taixe15a,sam:Milan16a}, we do a ($4$-fold) cross validation for all our experiments, except for the benchmark results in Section~\ref{subsec:experiments_benchmark_results}.

To assess the performance of the tracking algorithms we rely on standard MOT metrics, CLEAR MOT~\cite{sam:Bernadin08a} and MT/PT/ML~\cite{sam:Li09a}, which are also used by both benchmarks~\cite{sam:Geiger12a,sam:Leal-Taixe15a}.
This set of metrics measures recall and precision, both on a detection and trajectory level, counts the number of identity switches and fragmentations of trajectories and also provides an overall tracking accuracy (MOTA).

\subsection{Learned versus hand-crafted cost functions}
\label{subsec:experiments_learned_vs_crafted}
The main contribution of this paper is a novel way to automatically learn parameterized cost functions for a network flow based tracking model from data.
We illustrate the efficacy of the learned cost functions by comparing them with two standard choices for hand-crafted costs.
First, we follow \cite{sam:Leal-Taixe11a} and define $\costDet_\bboxIdx = \log(1 - \detProb{\det_\bboxIdx})$, where $\detProb{\det_\bboxIdx}$ is the detection probability, and
\begin{equation}
  \costLink_{\bboxIdx,\bboxIdxAlt} = -\log \textrm{E}\left( \frac{\|\detBox{\det_\bboxIdx} - \detBox{\det_\bboxIdxAlt}\|}{\Delta_t}, V_{\textrm{max}} \right) - \log(B^{\Delta_t - 1}),
  \label{eq:hc_cost_lealtaixe}
\end{equation}
where $\textrm{E}(V_t, V_{\textrm{max}}) = \frac{1}{2} + \frac{1}{2}\textrm{erf}(\frac{-V_t + 0.5 \cdot V_{\textrm{max}}}{0.25 \cdot V_{\textrm{max}}})$ with $\textrm{erf}(\cdot)$ being the Gauss error function and $\Delta_t$ is the frame difference between $\bboxIdx$ and $\bboxIdxAlt$.
While \cite{sam:Leal-Taixe11a} defines a slightly different network flow graph, we keep the graph definition the same (see Section~\ref{subsec:method_network_flow_formulation}) for all methods to ensure a fair comparison of the costs.
Second, we hand-craft our own cost function and define $\costDet_\bboxIdx = \alpha \cdot \detProb{\det_\bboxIdx}$ as well as
\begin{equation}
    \costLink_{\bboxIdx,\bboxIdxAlt} = (1 - \textrm{IoU}(\detBox{\det_\bboxIdx}, \detBox{\det_\bboxIdxAlt})) + \beta \cdot (\Delta_t - 1),
    \label{eq:hc_cost_ours}
\end{equation}
where $\textrm{IoU}(\cdot,\cdot)$ is the intersection over union.
We tune all parameters, \ie, $\costIn_\bboxIdx = \costOut_\bboxIdx = C$ (we did not observe any benefit when choosing these parameters separately), $B$, $V_{\textrm{max}}$, $\alpha$ and $\beta$, with grid search to maximize MOTA while balancing recall.
Note that the exponential growth of the search space \wrt the number of parameters makes grid search infeasible at some point.

With the same source of input information, \ie, bounding boxes $\bbox$ and detection confidences $\bboxP$, we train various types of parameterized functions with the algorithm proposed in Section~\ref{subsec:method_learning_cost_functions_from_data}.
For unary costs, we use the same parameterization as for the hand-crafted model, \ie, constants for $\costIn$ and $\costOut$ and a linear model for $\costDet$.
However, for the pair-wise costs, we evaluate a linear model, a one-layer MLP with $64$ hidden neurons and a two-layer MLP with $32$ hidden neurons in both layers.
The input feature $\costFin$ is the difference between the two bounding boxes, their detection confidences, the normalized time difference, as well as the IoU value.
We train all three models for $50$k iterations using ADAM~\cite{sam:Kingma15a} with a learning rate of $10^{-4}$, which we decrease by a factor of $10$ every $20$k iterations.

\begin{table}\footnotesize\centering
\begin{subtable}{1.0\columnwidth}
  \begin{tabular}{ l c c c c c c }
    \toprule
        & MOTA  & REC & PREC & MT & IDS & FRAG \\
    \midrule
Crafted~\cite{sam:Leal-Taixe11a} &           73.64 &           83.54 &           92.99 &           58.73 &             121 &             459 \\ 
                  Crafted-ours &           73.75 &           83.92 &           92.65 &           59.44 &              89 &             431 \\ 
\midrule
                  Linear &           73.51 &           83.47 &           92.99 &           59.08 &             132 &             430 \\ 
                  MLP 1 &           74.09 &           83.93 &           92.87 &           59.61 &              70 &             371 \\ 
             MLP 2 &           74.19 &           84.07 &           92.85 &           59.96 &              70 &             376 \\ 

    \bottomrule
  \end{tabular}
  \vspace{-0.15cm}
  \caption{}
  \vspace{0.1cm}
\end{subtable}
\begin{subtable}{1.0\columnwidth}
  \begin{tabular}{ l c c c c c c }
      \toprule
          & MOTA  & REC & PREC & MT & IDS & FRAG \\
      \midrule
Crafted~\cite{sam:Leal-Taixe11a} &           28.28 &           29.94 &           95.04 &            5.80 &             111 &            1063 \\ 
                  Crafted-ours &           29.19 &           34.01 &           87.88 &            6.77 &             142 &            1272 \\ 
\midrule
                  Linear &           28.25 &           38.01 &           80.09 &            9.67 &             342 &            1620 \\ 
                  MLP 1 &           31.05 &           37.51 &           85.81 &            8.32 &             282 &            1553 \\ 
             MLP 2 &           31.10 &           37.53 &           85.88 &            8.51 &             289 &            1562 \\ 

      \bottomrule
  \end{tabular}
  \vspace{-0.10cm}
  \caption{}
\end{subtable}
\vspace{-0.25cm}
\caption{
Learned vs. hand-crafted cost functions on a cross-validation on (a) KITTI-Tracking~\cite{sam:Geiger12a} and (b) MOT16~\cite{sam:Milan16a}.
}
\label{tbl:experiments_crafted_vs_learned}
\end{table}

Table~\ref{tbl:experiments_crafted_vs_learned} shows that our proposed training algorithm can successfully learn cost functions from data on both KITTI-Tracking and MOT16 data sets.
With the same input information given, our approach even slightly outperforms both hand-crafted baselines in terms of MOTA.
In particular, we observe lower identity switches and fragmentations on KITTI-Tracking and higher recall and mostly-tracked on MOT16.
While our hand-crafted function~\eqref{eq:hc_cost_ours} is inherently limited when objects move fast and IoU becomes $0$ (compared to~\eqref{eq:hc_cost_lealtaixe}~\cite{sam:Leal-Taixe11a}), both still achieve similar performance.
For both baselines, we did a hierarchical grid search to get good results.
However, an even finer grid search would be required to achieve further improvements.
The attraction of our method is that it obviates the need for such a tedious search and provides a principled way of finding good parameters.
We can also observe from the tables that non-linear functions (MLP 1 and MLP 2) perform better than linear functions (Linear), which is not possible in~\cite{sam:Wang15a}.

\subsection{Combining multiple input sources}
\label{subsec:experiments_combining_multiple_input_sources}
Recent works have shown that temporal and appearance features are often beneficial for MOT.
Choi~\cite{sam:Choi15a} presents a spatio-temporal feature (ALFD) to compare two detections, which summarizes statistics from tracked interest points in a $288$-dimensional histogram.
Leal-Taix\'{e}~\etal~\cite{sam:Leal-Taixe16a} show how to use raw RGB data with a Siamese network to compute an affinity metric for pedestrian tracking.
Incorporating such information into a tracking model typically requires (i) an isolated learning phase for the affinity metric and (ii) some hand-tuning to combine it with other affinity metrics and other costs in the model (\eg., $\costIn$, $\costDet$, $\costOut$).
In the following, we demonstrate the use of both motion and appearance features in our framework.

\textbf{Motion-features:}
In Table~\ref{tbl:experiments_combine_inputs}, we demonstrate the impact of the motion feature ALFD~\cite{sam:Choi15a} compared to purely spatial features on the KITTI-Tracking data set as in~\cite{sam:Choi15a}.
For each source of input, we compare both hand-crafted (C) and learned (L) pair-wise cost functions.
First, we use only the raw bounding box information (B), \ie, location and temporal difference and detection score.
For the hand-crafted baseline, we use the cost function defined in~\eqref{eq:hc_cost_lealtaixe}, \ie, \cite{sam:Leal-Taixe11a}.
Second, we add the IoU overlap (B+O) and use \eqref{eq:hc_cost_ours} for the hand-crafted baseline.
Third, we incorporate ALFD~\cite{sam:Choi15a} into the cost (B+O+M).
To build a hand-crafted baseline for (B+O+M), we construct a separate training set of ALFD features containing examples for positive and negative matches and train an SVM on the binary classification task.
During tracking, the normalized SVM scores $\hat{s}_{\textrm{A}}$ (a sigmoid function maps the raw SVM scores into $[0,1]$) are incorporated into the cost function
\begin{equation}
    \costLink_{\bboxIdx,\bboxIdxAlt} = (1-\textrm{IoU}(\detBox{\det_\bboxIdx}, \detBox{\det_\bboxIdxAlt})) + \beta \cdot (\Delta_t - 1) + \gamma \cdot (1 - \hat{s}_{\textrm{A}}),
    \label{eq:hc_cost_ours_alfd}
\end{equation}
where $\gamma$ is another hyper-parameter we also tune with grid-search.
For our learned cost functions, we use a $2$-layer MLP with $64$ neurons in each layer to predict $\costLink_{\bboxIdx,\bboxIdxAlt}$ for the (B) and (B+O) options.
For (B+O+M), we use a separate 2-layer MLP to process the $288$-dimensional ALFD feature, concatenate both $64$-dimensional hidden vectors of the second layers, and predict $\costLink_{\bboxIdx,\bboxIdxAlt}$ with a final linear layer.

Table~\ref{tbl:experiments_combine_inputs} again shows that learned cost functions outperform hand-crafted costs for all input sources, which is consistent with the previous experiment in Section~\ref{subsec:experiments_learned_vs_crafted}.
The table also demonstrates the ability of our approach to make effective use of the ALFD motion feature~\cite{sam:Choi15a}, especially for identity switches and fragmentations.
While it is typically tedious and suboptimal to combine such diverse features in hand-crafted cost functions, it is easy with our learning method because all parameters can still be jointly trained under the same loss function.

\begin{table}\footnotesize
\begin{center}
\begin{tabular}{ l c c c c c c }
    \toprule
    Inputs    & MOTA  & REC & PREC & MT & IDS & FRAG \\
    \midrule
                         (C) B &           73.64 &           83.54 &           92.99 &           58.73 &             121 &             459 \\ 
                         (L) B &           73.65 &           84.55 &           92.00 &           61.55 &              89 &             422 \\ 
\midrule
                       (C) B+O &           73.75 &           83.92 &           92.65 &           59.44 &              89 &             431 \\ 
                       (L) B+O &           74.12 &           84.13 &           92.69 &           60.49 &              55 &             361 \\ 
\midrule
                     (C) B+O+M &           73.07 &           85.07 &           90.92 &           61.73 &              43 &             386 \\ 
                     (L) B+O+M &           74.11 &           84.74 &           92.05 &           61.73 &              29 &             335 \\ 

    \bottomrule
\end{tabular}
\end{center}
\vspace{-0.5cm}
\caption{
We evaluate the influence of different types of input sources, raw detection inputs (B), bounding box overlaps (O) and the ALFD motion feature~\cite{sam:Choi15a} (M) for both learned (L) and hand-crafted (C) costs on KITTI-Tracking~\cite{sam:Geiger12a}.
}
\label{tbl:experiments_combine_inputs}
\end{table}

\begin{table}\footnotesize
\begin{center}
\begin{tabular}{ l c c c c c c }
    \toprule
    Unary cost    & MOTA  & REC & PREC & MT & IDS & FRAG \\
    \midrule
Crafted~\cite{sam:Leal-Taixe11a} &           30.55 &           38.54 &           83.70 &           11.60 &             194 &             853 \\
                  Crafted-ours &           30.43 &           38.98 &           82.69 &           11.40 &             156 &             825 \\
                         (B+O) &           28.94 &           43.63 &           75.47 &           14.00 &             204 &             962 \\
                      Au+(B+O) &           39.08 &           46.99 &           86.71 &           15.60 &             285 &            1062 \\
                   Au+(B+O+Ap) &           39.23 &           47.17 &           86.50 &           15.80 &             233 &             954 \\

    \bottomrule
\end{tabular}
\end{center}
\vspace{-0.5cm}
\caption{
Using appearance for unary (Au) and pair-wise (Ap) cost functions clearly improves tracking performance.
}
\label{tbl:experiments_rgb_patch_unary}
\end{table}

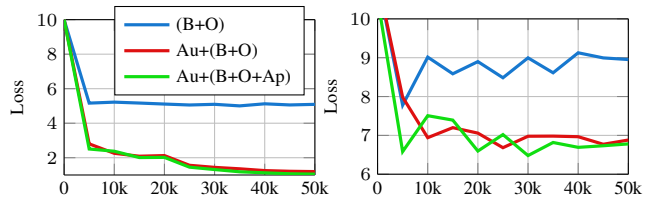
\begin{figure}\centering
\begin{subfigure}{.5\columnwidth}\centering
  \setlength{\tikzfigwidth}{0.80\columnwidth}
  \setlength{\tikzfigheight}{0.50\columnwidth}
  \scriptsize
  \hspace{-0.5cm}
\begin{tikzpicture}

\definecolor{myorange}{RGB}{252, 120, 23} 
\definecolor{myblue}{RGB}{23, 120, 207} 
\definecolor{mygreen}{RGB}{16, 222, 17} 
\definecolor{myred}{RGB}{222, 16, 17} 

\begin{axis}[
width=\tikzfigwidth,
height=0.991935\tikzfigheight,
at={(0\tikzfigwidth,0\tikzfigheight)},
scale only axis,
separate axis lines,
every outer x axis line/.append style={black},
every x tick label/.append style={font=\color{black}},
xmin=0,
xmax=5,
xtick={0,1,2,3,4,5},
xticklabels={0,10k,20k,30k,40k,50k},
xmajorgrids,
every outer y axis line/.append style={black},
every y tick label/.append style={font=\color{black}},
ymin=1.0,
ymax=10.0,
ylabel={Loss},
ymajorgrids,
legend style={legend cell align=left,align=left,draw=black,at={(0.20,0.8)},anchor=west}
]
\addplot[myblue,solid,very thick]
  coordinates{
    (0.0, 10.00000000)
    (0.5,  5.16438532)
    (1.0,  5.22028875)
    (1.5,  5.16814137)
    (2.0,  5.11017561)
    (2.5,  5.05595112)
    (3.0,  5.09502220)
    (3.5,  5.00397825)
    (4.0,  5.12424564)
    (4.5,  5.05856133)
    (5.0,  5.09283924)
};
\addlegendentry{(B+O)};
\addplot[myred,solid,very thick]
  coordinates{
    (0.0, 10.00000000)
    (0.5,  2.80424047)
    (1.0,  2.25608397)
    (1.5,  2.08861661)
    (2.0,  2.12521791)
    (2.5,  1.56205344)
    (3.0,  1.44181323)
    (3.5,  1.35867107)
    (4.0,  1.25517666)
    (4.5,  1.21531332)
    (5.0,  1.19704294)
};
\addlegendentry{Au+(B+O)};
\addplot[mygreen,solid,very thick]
  coordinates{
    (0.0, 10.00000000)
    (0.5,  2.50603914)
    (1.0,  2.37656760)
    (1.5,  2.01027560)
    (2.0,  2.02061415)
    (2.5,  1.45334256)
    (3.0,  1.31562972)
    (3.5,  1.18518567)
    (4.0,  1.12588966)
    (4.5,  1.06955934)
    (5.0,  1.05302429)
};
\addlegendentry{Au+(B+O+Ap)};
\end{axis}


\end{tikzpicture}

\end{subfigure}%
\begin{subfigure}{.5\columnwidth}\centering
  \setlength{\tikzfigwidth}{0.80\columnwidth}
  \setlength{\tikzfigheight}{0.50\columnwidth}
  \scriptsize
  \hspace{-0.5cm}
\begin{tikzpicture}\scriptsize

\definecolor{myorange}{RGB}{252, 120, 23} 
\definecolor{myblue}{RGB}{23, 120, 207} 
\definecolor{mygreen}{RGB}{16, 222, 17} 
\definecolor{myred}{RGB}{222, 16, 17} 

\begin{axis}[
width=\tikzfigwidth,
height=0.991935\tikzfigheight,
at={(0\tikzfigwidth,0\tikzfigheight)},
scale only axis,
separate axis lines,
every outer x axis line/.append style={black},
every x tick label/.append style={font=\color{black}},
xmin=0,
xmax=5,
xtick={0,1,2,3,4,5},
xticklabels={0,10k,20k,30k,40k,50k},
xmajorgrids,
every outer y axis line/.append style={black},
every y tick label/.append style={font=\color{black}},
ymin=6.0,
ymax=10.0,
ylabel={Loss},
ymajorgrids
]
\addplot[myblue,solid,very thick]
  coordinates{
    (0.0, 11.01856079)
    (0.5,  7.77829838)
    (1.0,  9.01510429)
    (1.5,  8.58585072)
    (2.0,  8.89965248)
    (2.5,  8.48376751)
    (3.0,  8.99652767)
    (3.5,  8.61180592)
    (4.0,  9.12664890)
    (4.5,  8.99522591)
    (5.0,  8.95303822)
};
\addplot[myred,solid,very thick]
  coordinates{
    (0.0, 11.02442780)
    (0.5,  7.98415756)
    (1.0,  6.94077682)
    (1.5,  7.19915342)
    (2.0,  7.05857229)
    (2.5,  6.68313837)
    (3.0,  6.97806025)
    (3.5,  6.98116302)
    (4.0,  6.96582031)
    (4.5,  6.77122402)
    (5.0,  6.88053417)
};
\addplot[mygreen,solid,very thick]
  coordinates{
    (0.0, 10.52216721)
    (0.5,  6.58335495)
    (1.0,  7.50703144)
    (1.5,  7.39329433)
    (2.0,  6.59696198)
    (2.5,  7.02059460)
    (3.0,  6.48031712)
    (3.5,  6.81714439)
    (4.0,  6.69205761)
    (4.5,  6.73279095)
    (5.0,  6.78057766)
};
\end{axis}


\end{tikzpicture}

\end{subfigure}
\caption{
The difference in the loss on the training (left) and validation set (right) over 50k iterations of training for models w/ (Au,Ap) and w/o appearance features.
}
\label{fig:loss_appearance_input}
\end{figure}

\textbf{Appearance features:}
Here, we investigate the impact of raw RGB data on both unary and pair-wise costs of the network flow formulation.
We use the MOT15 data set~\cite{sam:Leal-Taixe15a} and the provided ACF detections~\cite{sam:Dollar14a}.
First, we integrate the raw RGB data into the unary cost $\costDet_\bboxIdx$ (Au).
For each detected bounding box $\detBox{\det_\bboxIdx}$, we crop the underlying RGB patch $I_\bboxIdx$ with a fixed aspect ratio, resize the patch to $128 \times 64$ and define the cost
\begin{equation}
    \costDet_\bboxIdx = \cost_{\textrm{conf}}(\detProb{\det_\bboxIdx}; \costFpa_{\textrm{conf}}) + \cost_{\textrm{Au}}(I_\bboxIdx; \costFpa_{\textrm{Au}}),
    \label{eq:cost_unary_rgbpatch}
\end{equation}
which consists of one linear function taking the detection confidence and one deep network taking the image patch.
We choose ResNet-50~\cite{sam:He16a} to extract features for $\cost_{\textrm{Au}}$ but any other differentiable function can be used as well.

Second, we use a Siamese network (same as for unary term) that compares RGB patches of two detections, similar to~\cite{sam:Leal-Taixe16a} but without optical flow information.
As with the motion features above, we use a two-stream network to combine spatial information (B+O) with appearance features (Ap).
The hidden feature vector of a 2-layer MLP (B+O) is concatenated with the difference of the hidden features from the Siamese network.
A final linear layer predicts the costs $\costLink_{\bboxIdx,\bboxIdxAlt}$ of the pair-wise terms.

Table~\ref{tbl:experiments_rgb_patch_unary} shows that integrating RGB information into the detection cost Au+(B+O) improves tracking performance significantly over the baselines.
Using the RGB information in the pair-wise cost as well Au+(B+O+Ap) further improves results, especially for identity switches and fragmentations.
Figure~\ref{fig:loss_appearance_input} visualizes the loss on the training and validation set for the three learning-based methods, which again shows the impact of appearance features.
Note, however, that the improvement is limited because we still rely on the underlying ACF detector and are not able to improve recall over the recall of the detector.
But the experiment clearly shows the potential ability to integrate deep network based object detectors directly into an end-to-end tracking framework.
We plan to investigate this avenue in future work.

\subsection{Weighting the loss function}
\label{subsec:experiments_weighting_the_loss_function}
For completeness, we also investigate the impact of different weighting schemes for the loss function defined in Section~\ref{subsec:method_learning_cost_functions_from_data_gt_and_loss}.
First, we compare our loss function without any weighting (none) with the loss defined in \cite{sam:Wang15a}.
We also do this for an $\ell_1$ loss.
We can see from the first part in Table~\ref{tbl:experiments_loss_weighting} that both achieve similar results but \cite{sam:Wang15a} achieves slightly better identity switches and fragmentations.
By decreasing $\omega_{\textrm{basic}}$ we limit the impact of ambiguous cases (see Section~\ref{subsec:method_learning_cost_functions_from_data_gt_and_loss}) and can observe a slight increase in recall and mostly tracked.
Also, we can influence the trade-off between precision and recall with $\omega_{\textrm{pr}}$ and we can lower the number of identity switches by increasing $\omega_{\textrm{links}}$.

\begin{table}\footnotesize
\begin{center}
\begin{tabular}{ l c c c c c c }
    \toprule
    Weighting    & MOTA  & REC & PREC & MT & IDS & FRAG \\
    \midrule
                          none &           74.07 &           82.84 &           93.78 &           57.67 &              53 &             333 \\ 
            \cite{sam:Wang15a} &           73.99 &           82.90 &           93.63 &           57.32 &              43 &             331 \\ 
                none-$\ell_1$ &           73.90 &           83.43 &           93.17 &           58.73 &              77 &             362 \\ 
  \cite{sam:Wang15a}-$\ell_1$ &           73.92 &           83.19 &           93.38 &           58.73 &              71 &             357 \\ 
\midrule
              $\omega_{\textrm{basic}} = 0.1$ &           74.15 &           84.11 &           92.72 &           60.49 &              51 &             360 \\ 
              $\omega_{\textrm{basic}} = 0.5$ &           74.13 &           83.90 &           92.92 &           59.96 &              62 &             363 \\ 
\midrule
                 $\omega_{\textrm{pr}} = 0.3$ &           66.84 &           70.68 &           98.35 &           28.92 &              34 &             216 \\ 
                 $\omega_{\textrm{pr}} = 1.5$ &           73.28 &           85.52 &           90.85 &           63.49 &              80 &             387 \\ 
\midrule
              $\omega_{\textrm{links}} = 1.5$ &           74.14 &           84.53 &           92.31 &           61.38 &              45 &             357 \\ 
              $\omega_{\textrm{links}} = 2.0$ &           74.10 &           84.80 &           92.03 &           61.38 &              42 &             358 \\ 

    \bottomrule
\end{tabular}
\end{center}
\vspace{-0.5cm}
\caption{
Differently weighting the loss function provides a trade-off between various behaviors of the learned costs.
}
\label{tbl:experiments_loss_weighting}
\end{table}

\subsection{Benchmark results}
\label{subsec:experiments_benchmark_results}
Finally, we evaluate our learned cost functions on the benchmark test sets.
For KITTI-Tracking~\cite{sam:Geiger12a}, we train cost functions equal to the ones described in Section~\ref{subsec:experiments_combining_multiple_input_sources} with ALFD motion features~\cite{sam:Choi15a}, \ie, (B+O+M) in Table~\ref{tbl:experiments_combine_inputs}.
We train the models on the full training set and upload the results on the benchmark server.
Table~\ref{tbl:experiments_benchmark_kitti} compares our method with other off-line approaches that use RegionLet detections~\cite{sam:Wang13a}.
While~\cite{sam:Choi15a} achieves better results on the benchmark, their approach includes a complex graphical model and a temporal model for trajectories.
The fair comparison is with Wang and Fowlkes~\cite{sam:Wang16a}, which is the most similar approach to ours.
While we achieve better MOTA, it is important to note that the comparison needs to be taken with a grain of salt.
We include motion features in the form of ALFD~\cite{sam:Choi15a}.
On the other hand, the graph in~\cite{sam:Wang16a} is more complex as it also accounts for trajectory interactions.

We also evaluate on the MOT15 data set~\cite{sam:Leal-Taixe15a}, where we choose the model that integrates raw RGB data into the unary costs, \ie, Au+(B+O) in Table~\ref{tbl:experiments_rgb_patch_unary}.
We achieve an MOTA value of $26.8$, compared to $25.2$ for~\cite{sam:Wang16a} (most similar model) and $29.0$ for~\cite{sam:Leal-Taixe16a} (using RGB data for pair-wise term).
We again note that~\cite{sam:Leal-Taixe16a} additionally integrates optical flow into the pair-wise term.
The impact of RGB features is not as pronounced as in our cross-validation experiment in Table~\ref{tbl:experiments_rgb_patch_unary}.
The most likely reason we found for this scenario is over-fitting of the unary terms.

\begin{figure}\centering
  \input{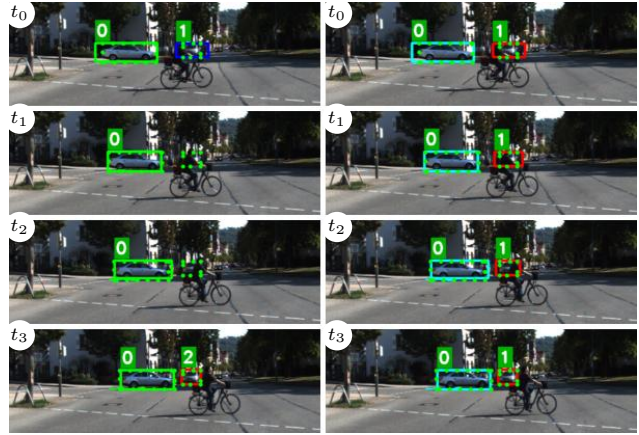}
  \vspace{-0.1cm}
  \caption{
    A qualitative example showing a failure case of the hand-crafted costs (left) compared to the learned costs (right), which leads to a fragmentation.
    The green dotted boxes are ground truth, the solid colored are ones tracked objects.
    The numbers are the object IDs.
    Best viewed in color and zoomed.
  }
  \label{fig:experiments_qualitative_example}
\end{figure}

Figure~\ref{fig:experiments_qualitative_example} also gives a qualitative comparison between hand-crafted and learned cost functions on KITTI~\cite{sam:Geiger12a}.
The supplemental material contains more qualitative results.

\begin{table}\footnotesize
\begin{center}
\begin{tabular}{l c c c c c c}
    \toprule
    Method                & MOTA & MOTP & MT & ML & IDS & FRAG \\
    \midrule
    \cite{sam:Lenz15a}          & 60.84 & 78.55 & 53.81 & 7.93  & 191 & 966  \\
    \cite{sam:Choi15a}          & 69.73 & 79.46 & 56.25 & 12.96 & 36  & 225  \\
    \cite{sam:Milan13a}         & 55.49 & 78.85 & 36.74 & 14.02 & 323 & 984  \\
    \cite{sam:Wang16a}          & 66.35 & 77.80 & 55.95 & 8.23  & 63  & 558  \\
    Ours                        & 67.36 & 78.79 & 53.81 & 9.45  & 65  & 574  \\
    \bottomrule
\end{tabular}
\end{center}
\vspace{-0.5cm}
\caption{
Results on KITTI-Tracking~\cite{sam:Geiger12a} from 11/04/16.
}
\label{tbl:experiments_benchmark_kitti}
\end{table}

\section{Conclusion}
\label{sec:conclusion}
Our work demonstrates how to learn a parameterized cost function of a network flow problem for multi-object tracking in an end-to-end fashion.
The main benefit is the gained flexibility in the design of the cost function.
We only assume it to be parameterized and differentiable, enabling the use of powerful neural network architectures.
Our formulation learns the costs of \emph{all} variables in the network flow graph, avoiding the delicate task of hand-crafting these costs.
Moreover, our approach also allows for easily combining different sources of input data.
Evaluations on three public data sets confirm these benefits empirically.

For future works, we plan to integrate object detectors end-to-end into this tracking model, investigate more complex network flow graphs with trajectory interactions and explore applications to max-flow problems.


{\small
\bibliographystyle{ieee}
\bibliography{myshortstrings,sam_schulter_2016_01_01}
}

\end{document}